\documentclass[10pt,twocolumn,letterpaper]{article}

\usepackage{multirow}
\usepackage[table ]{xcolor}
\usepackage[pagenumbers]{cvpr} 

\PassOptionsToPackage{table,dvipsnames}{xcolor}
\usepackage[dvipsnames]{xcolor}

\definecolor{cvprblue}{rgb}{0.21,0.49,0.74}
\usepackage[pagebackref,breaklinks,colorlinks,citecolor=cvprblue]{hyperref}

\def\paperID{11556} 
\def\confName{CVPR}
\def\confYear{2024}

\title{STAF: 3D Human Mesh Recovery from Video with Spatio-Temporal Alignment Fusion}

\author{Wei Yao$^1$, Hongwen Zhang$^2$, Yunlian Sun$^1$, Jinhui Tang$^1$\\
$^1$Nanjing University of Science and Technology, $^2$Beijing Normal University
}

\begin{document}
\maketitle
\begin{abstract}
The recovery of 3D human mesh from monocular images has significantly been developed in recent years. However, existing models usually ignore spatial and temporal information, which might lead to mesh and image misalignment and temporal discontinuity. For this reason, we propose a novel Spatio-Temporal Alignment Fusion (STAF) model. As a video-based model, it leverages coherence clues from human motion by an attention-based Temporal Coherence Fusion Module (TCFM). As for spatial mesh-alignment evidence, we extract fine-grained local information through predicted mesh projection on the feature maps. Based on the spatial features, we further introduce a multi-stage adjacent Spatial Alignment Fusion Module (SAFM) to enhance the feature representation of the target frame. In addition to the above, we propose an Average Pooling Module (APM) to allow the model to focus on the entire input sequence rather than just the target frame. This method can remarkably improve the smoothness of recovery results from video. Extensive experiments on 3DPW, MPII3D, and H36M demonstrate the superiority of STAF. We achieve a state-of-the-art trade-off between precision and smoothness. Our code and more video results are on the project page \href{https://yw0208.github.io/staf/}{https://yw0208.github.io/staf/}.
\end{abstract}    
\section{Introduction}
{A}{s} a promising technology, video-based human mesh recovery can be used for many tasks such as motion monitoring, virtual try-on, VR, etc. It also contributes to traditional human-centered computer vision research, such as action recognition~\cite{humanaction} and pose estimation~\cite{poseestimation,mutipose,Anatomy-Aware}. Therefore, it has received wide attention from the research community and has been developed rapidly in recent years~\cite{tian2022survey}. Especially after the emergence of parametric models that can describe the human body surface in detail (e.g., SMPL~\cite{loper2015smpl}), many excellent models have emerged and achieved good results with the development of deep learning. 

Recovering the 3D human body from a video is a more complex problem than recovering it from a single image. Many video-based works tried to find effective methods to obtain temporal information. Currently, there are mainly convolutional neural network (CNN) and recurrent neural network (RNN) for learning temporal information~\cite{tcmr}, \cite{doersch2019sim2real}, \cite{kanazawa2019learning}, \cite{vibe}, \cite{meva}. It should be noted that both CNN and RNN are better at learning local information~\cite{vaswani2017attention}, \cite{nonlocal} but have difficulty when handling long-range temporal dependencies. Therefore, finding a simple and efficient mechanism for acquiring temporal information is necessary. To leverage temporal cues, the mainstream methods simply fuse the global features extracted from ResNet~\cite{resnet} or HRNet~\cite{hrnet} and then use this feature to get the final result. According to previous works~\cite{hrnet}, \cite{unet}, \cite{newell2016stacked}, \cite{lin2017feature}, \cite{bmp}, feature map tends to retain high-level information after reducing the spatial dimension while ignoring spatial information as well as local details. There are many studies attempting to solve this challenge using pixel-level information, such as body part segmentation~\cite{nbf}, \cite{hund}, \cite{thundr}, UV map~\cite{pymaf}, \cite{xu2019denserac}, \cite{danet}, \cite{pose2uv} and optical flow~\cite{alldieck2017optical}, \cite{arnab2019exploiting}, \cite{dstvibe}. But these usually make the model too bloated and still challenging to learn the body structure prior and local details. Moreover, existing video-based and image-based models typically showed severe jitter when applied to video. And this jitter phenomenon cannot be effectively mitigated with the increase in recovery precision. Although there are previous works that attempted to solve this problem, they all sacrifice the recovery precision to some extent. So, achieving a better balance between precision and smoothness is still a difficult challenge.

\begin{figure*}[h]
    \centering
    \includegraphics[width=\textwidth]{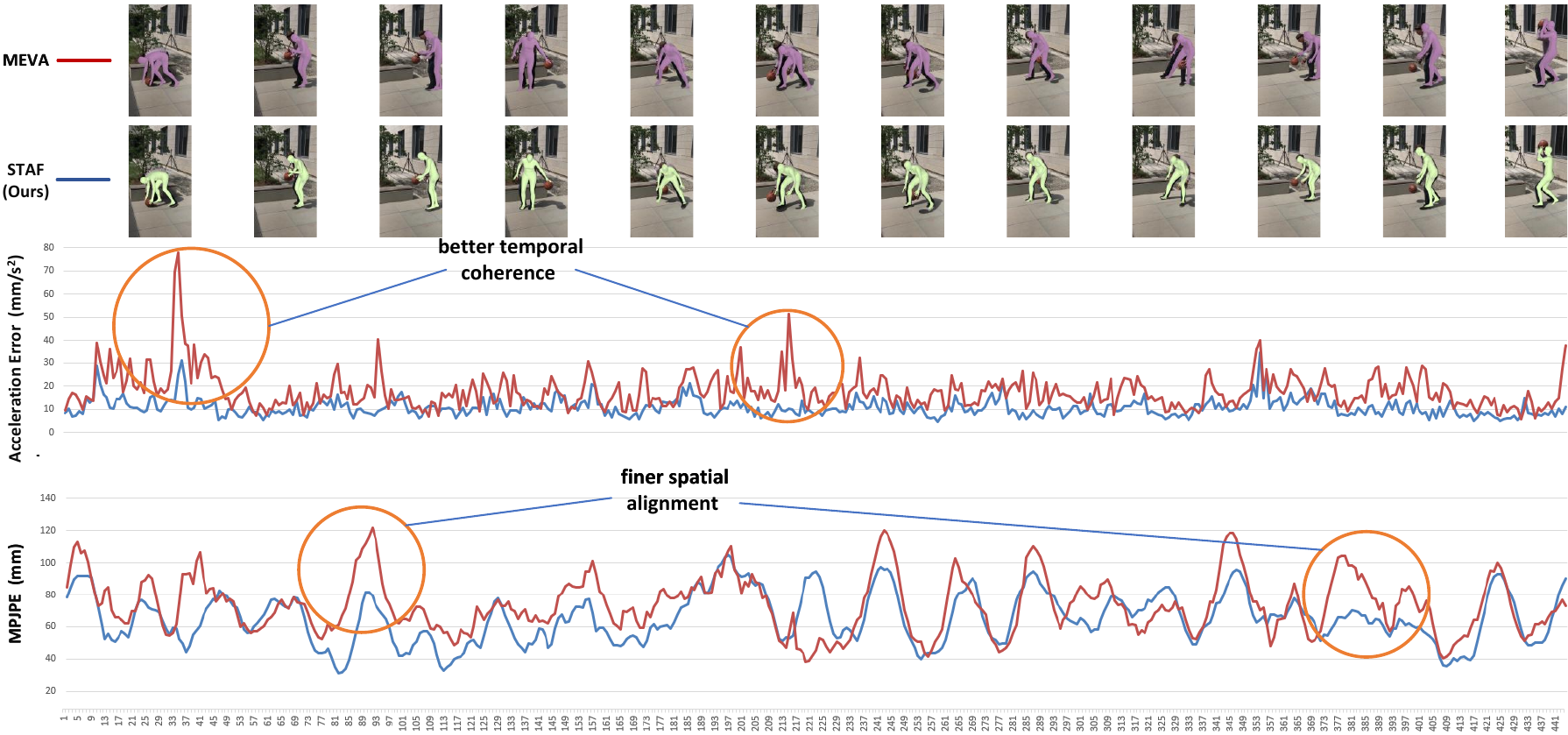}
    \caption{Comparison with traditional video-based model MEVA~\cite{meva}. We choose MPJPE and acceleration error to measure the model’s performance in space and time. Thanks to our spatio-temporal fusion mechanism, our STAF surpasses MEVA in both metrics.}
    \label{fig:accel}
\end{figure*}

To address these issues, we propose a spatio-temporal alignment fusion (STAF) model for recovering 3D human meshes from videos. In STAF, a feature pyramid is introduced into the video domain for 3D human reconstruction as the backbone to preserve the original information to the maximum extent. Based on this, we propose a temporal coherence fusion module (TCFM), a spatial alignment fusion module (SAFM) and an average pooling module (APM) for the three problems. In this way, STAF can fully utilize the spatio-temporal information of the input image sequence and achieve a breakthrough in both precision and smoothness with the support of APM. As shown in Fig. \ref{fig:accel}, STAF outperforms the previous SOTA method in both terms of precision and smoothness.

Specifically, TCFM no longer uses global features as input. Instead, we collect features as input by grid projection on high-dimensional spatial features. This method preserves the original spatial position information to a large extent. In the 3D human reconstruction task, the so-called temporal information refers more to the consistency of human shape and the continuity of pose changes. Therefore, it is necessary to retain the original spatial position information for better learning of the temporal information. When choosing which network architecture to use for temporal encoding, we adopt a self-attention mechanism that is better at establishing long-range dependencies. However, the traditional self-attention module encodes the features before calculating the attention weights. As shown in $M_{con}$ of Fig \ref{fig:moca}, we find that this process could destroy the original feature space and instead make it difficult to establish the correct temporal dependencies. For this reason, we add the other self-similarity matrix $M_{sim}$, which can guide TCFM to encode the temporal information better and thus get more accurate initial human meshes. These initial human meshes enable SAFM to obtain better spatial information about the human body in the following step.

\begin{figure*}[h]
    \centering
    \includegraphics[width=0.99\textwidth]{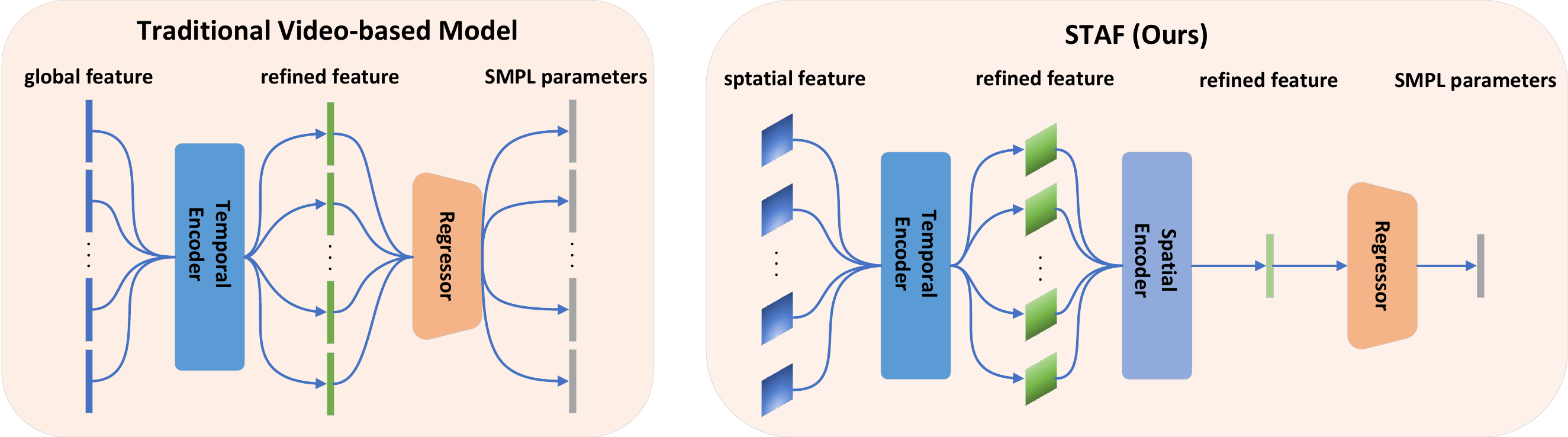}
    \caption{The difference between traditional video-based models and our STAF.  STAF has an additional spatial encoder compared to traditional video-based models. As a result, STAF can obtain more comprehensive refined features and achieve higher recovery precision.}
    \label{fig:seq2frame}
\end{figure*}

As shown in Fig \ref{fig:seq2frame}, compared to traditional models, STAF goes beyond the fusion of temporal information, and further incorporates spatial information. There are two crucial points about SAFM: the extraction of human spatial features, and the other is how to enhance the feature representation of the target frame. We use the projection of the initial human meshes on the feature maps to obtain human spatial features. This has two advantages. First, the mesh alignment cues can be used to correct the result parameters effectively. More importantly, since the features are extracted only in the human body region of the feature map, the model can obtain richer semantic information and focus more on informative human areas by reducing interference from the background. After getting the human spatial features, we need to use them to enhance the feature representation of the target frame. Considering that adjacent images' human shape and pose are more similar, we adopt a multi-stage attention-based adjacent feature fusion mechanism, as shown in Fig \ref{fig:hafi}. The human spatial information enables STAF to obtain a more precise recovery mesh of the target frame.

But as mentioned earlier, like other traditional models, even though STAF utilizes spatio-temporal information to improve the accuracy further, the smoothness is still not sufficiently improved. The reason for this is that the model cannot take into account the whole input sequence but focuses only on improving the recovery precision of the target frame. This leads to a lack of transition from frame to frame, which eventually causes frequent and noticeable jitter in the recovered human body. For this reason, we propose the APM that allows the model to focus on the entire input sequence, using each frame’s information to generate results that match the human motion in the sequence. This module can significantly improve the smoothness without affecting accuracy. It is worth mentioning that it is also applicable to many existing models. At last, we summarize our contributions as follows:
\begin{itemize}
\item[$\bullet$] For the first time, multi-scale spatial features are introduced into the 3D human mesh recovery task in the video domain. We propose a novel spatio-temporal alignment fusion model to exploit both spatial and temporal information. We propose an effective spatio-temporal feature interaction and integration mechanism that enables the model to take full advantage of motion continuity cues and human spatial information to recover more precise 3D human mesh.
\item[$\bullet$] We find an effective method to significantly improve the smoothness of the estimated mesh sequences from the video. We find that the main reason for the discontinuity of recovered human motion is that traditional models usually focus only on the target frame but not the overall sequence. With our proposed APM, we achieve a remarkable reduction in acceleration error and demonstrate experimentally that the method is somewhat generalizable.
\item[$\bullet$] Extensive experiments on three standard benchmark datasets show that STAF achieves state-of-the-art performance with a better trade-off between precision and smoothness. 
\end{itemize}
\section{Related Work}
\subsection{Image-based 3D Human Mesh Recovery}
Research on 3D human reconstruction started early and saw explosive growth after the emergence of human parametric models~\cite{anguelov2005scape}, \cite{loper2015smpl}, \cite{xu2020ghum} and human datasets with 3D labels~\cite{human36m}, \cite{3dpw}, \cite{MPII3d}. The first works in this field were based on optimization. These optimization-based methods let the parametric model constantly fit the obtained 2D labels (including silhouettes, 2D joint points, part segmentation, etc.) together with the human pre-existing prior~\cite{sigal2007combined}, \cite{guan2009estimating}, \cite{bogo2016keep}. In 2018, Angjoo et al. proposed the HMR model~\cite{hmr}, which was the first end-to-end regression-based model with a single monocular image as input. Using ResNet50~\cite{resnet} to extract features, HMR used an Iterative Error Feedback (IEF) loop regressor to get the final result and further adopted an action discriminator to ensure the reasonableness of the output 3D reconstruction. Since regression-based models have an absolute speed advantage as well as broader applicability than optimization-based models, a large number of excellent regression-based models~\cite{kolotouros2019convolutional}, \cite{yao2019densebody}, \cite{pymaf}, \cite{zhang2023pymaf}, \cite{cliff}, \cite{lin2021mesh}, \cite{kolotouros2021probabilistic}, \cite{sun2021monocular}, \cite{PQ-GCN} have emerged since then. However, image-based methods have their inherent limitations. Even compared to the latest PQ-GCN\cite{PQ-GCN}, which was carefully designed, our video-based STAF not only exceeds it in terms of accuracy but also offers much better smoothness.

While regression-based models have proliferated, optimization-based models have not fallen out of favor. Instead, they are combined with regression-based models to obtain models that can generate more accurate human body mesh~\cite{spin}, \cite{eft}. Such models generally used regression-based models to generate better initial results and then used the optimization process to obtain more accurate results. Researchers usually use such models to add 3D pseudo-labels to 2D training datasets, which can significantly facilitate the training of their models. 

\subsection{Video-based 3D Human Mesh Recovery}
In terms of practical applications, the application of 3D human reconstruction will be more based on videos, and the continuity of human motion contains rich temporal information that can be used. As a result, several video-based models have emerged in recent years. Currently, there are two main categories: sequence-to-sequence~\cite{vibe}, \cite{maed}, \cite{meva}, \cite{doersch2019sim2real}, \cite{arnab2019exploiting}, where multiple images are input andes all the corresponding human meshes are output, and sequence-to-single-frame~\cite{kanazawa2019learning}, \cite{dsd}, \cite{tcmr}, \cite{wei2022capturing}, where multiple images are input but only the result of the target frame is output. Earlier, there was Arnab et al. ~\cite{arnab2019exploiting}, which used the entire video as input, generated initial results using an off-the-shelf 2D joint keypoint detector~\cite{papandreou2017towards} and a 3D human reconstruction model~\cite{hmr}, and then continuously optimized the results using temporal coherence. There are also methods extracting features that allow models to learn temporal information adaptively. For example, HMMR~\cite{kanazawa2019learning} used full convolutional networks to encode temporal information, and MEVA~\cite{meva} adopted recurrent neural networks to learn. Among them, a classic work is VIBE~\cite{vibe}, which added a GRU-based module to encode temporal information based on HMR~\cite{hmr} and further designed a temporal version of motion discriminator to ensure the rationality of the output human mesh. With the rise of Transformer~\cite{vaswani2017attention}, the model MAED \cite{maed}, which used an attention mechanism to learn the continuity of each joint movement, achieved excellent performance.
\begin{figure*}[h]
    \centering
    \includegraphics[width=0.99\textwidth]{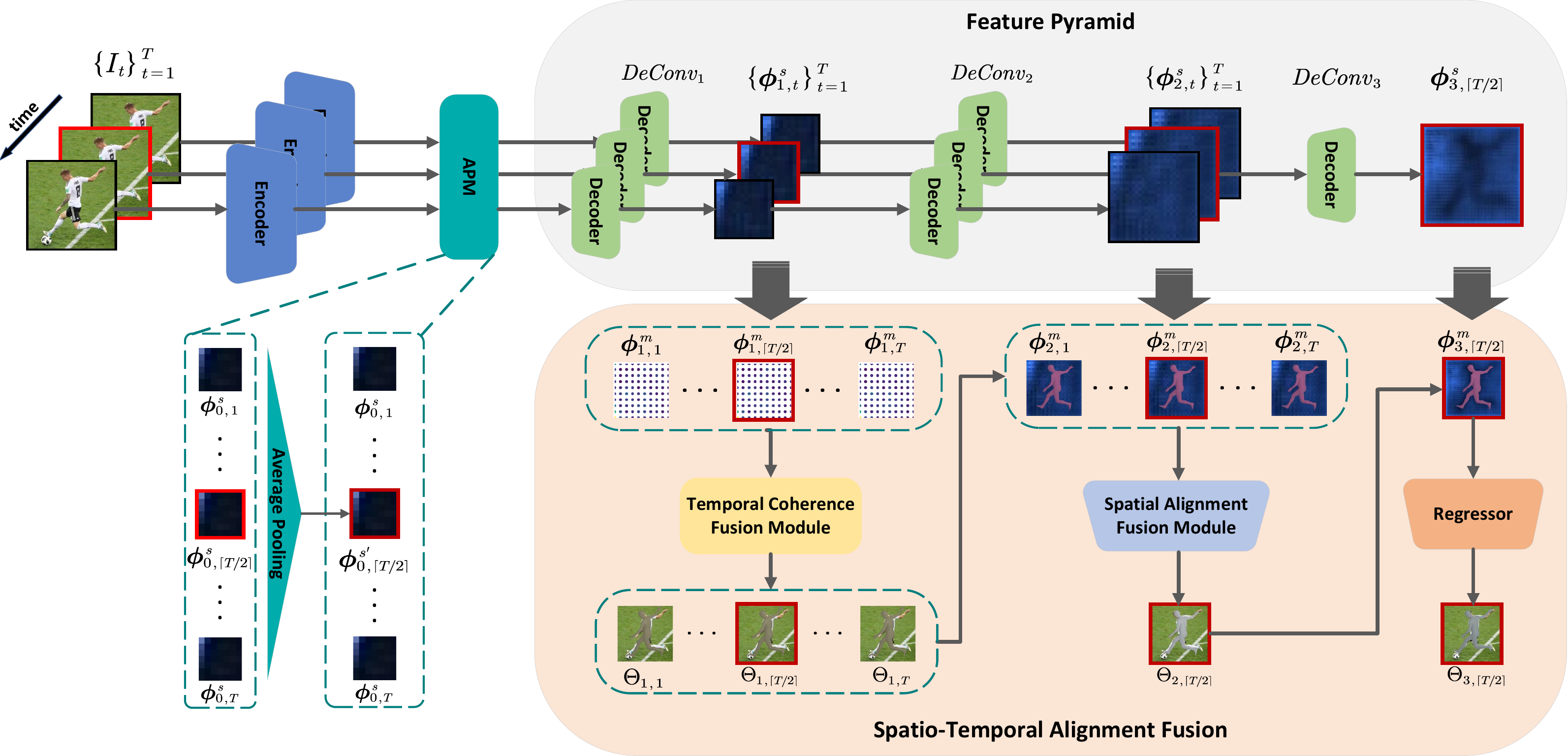}
    \caption{The overall framework of STAF. We input $T$ images and output the reconstruction result of the target frame $I_{\lceil T/2 \rceil}$ with a red border. We employ a feature pyramid to retain multi-scale spatial information and use projection down-sampling to obtain fine-grained local information. Also, to make full use of the spatio-temporal information, we add an average pooling module, a temporal coherence fusion module and a spatial alignment fusion module. The temporal coherence fusion module is described in Sec \ref{moca}, and the spatial alignment fusion module is in Sec \ref{hafi}. Please refer to Sec \ref{STAF} for the entire process of our method.}
    \label{fig:mymodel}
\end{figure*}
\subsection{Temporal Continuity of 3D Human Mesh Recovery}
When 3D human reconstruction is transferred from a single image to a video, it is not enough to emphasize only the accuracy of the reconstructed mesh. In fact, the visual discomfort caused by the incoherence of human motion is even more pronounced than the inaccuracy. Since the acceleration error proposed by HMMR~\cite{kanazawa2019learning} to measure the smoothness of the recovery results, there have been many works~\cite{vibe}, \cite{maed}, \cite{meva}, \cite{tcmr}, \cite{wei2022capturing} adopted this measure. Theoretically, the more accurate the human reconstruction results are, the lower the acceleration error and the smoother the estimated human motion. However, in practice, with the current accuracy, it is not yet possible to significantly reduce the acceleration error by increasing accuracy. Let us see two structurally similar models, MEVA~\cite{meva} and VIBE~\cite{vibe}. MEVA sacrificed reconstruction accuracy to improve smoothness, and VIBE improved accuracy but dramatically increased acceleration error. As far as the latest work is concerned, MAED~\cite{maed} improved the recovery accuracy to a very high level, but the fluency was much poorer than TCMR~\cite{tcmr} and MPS-Net~\cite{wei2022capturing}. These two works improved the smoothness to an unprecedented level without reducing accuracy. On the one hand, TCMR provided a method to remove the residual connections of features and reduce the feature dependence on the current frame. On the other hand, MPS-Net experimentally demonstrated that its feature integration module named HAFI could significantly reduce acceleration error. Inspired by the above two works, we go a step further and propose a more straightforward method to reduce the dependence on the target frame and significantly improve the smoothness without compromising accuracy.
\section{Method}

The whole framework of STAF is shown in Fig \ref{fig:mymodel}. With features extracted from input images, we first go through APM to weaken the influence of the target frame but strengthen the model's dependence on the whole sequence. After that, TCFM is designed to learn the temporal information to get initial human meshes. With these initial body meshes, we can obtain finer spatial alignment cues. Next, we propose SAFM to fully integrate these cues to strengthen the target frame's body spatial representation and further correct its recovery result. Finally, the fine-grained local information is extracted by projection sampling and fed into the regressor to obtain the final result. In this section, we present the details of STAF. We first introduce some basic knowledge, including the SMPL model and feature sampling. Then we show two crucial submodules, TCFM and SAFM, and summarize the whole framework at last.

\subsection{3D Human Representation}
In this work, a parametric model called SMPL~\cite{loper2015smpl} is used to encode the 3D surface of the human body, which is one of the most widely used 3D human models. In total, the SMPL model parameters $\Theta$ consist of three parts: shape $\boldsymbol{\beta }$, pose $\boldsymbol{\theta }$, and camera $\boldsymbol{\pi }$. The shape parameters $\boldsymbol{\beta } \in \mathbb{R} ^{10}$ consist of the first 10 coefficients of the PCA shape space, including the body weight, height, and the proportion of each limb. The pose parameters $\boldsymbol{\theta } \in \mathbb{R} ^{3J}$ use the 3D rotation of each joint point relative to its parent joint to describe the pose of the human body, where $J=23$. After obtaining $\boldsymbol{\theta }$ and $\boldsymbol{\beta }$, we can input them to a pre-trained function to obtain $M\left( \boldsymbol{\theta } ,\boldsymbol{\beta } \right) \in \mathbb{R} ^{3\times N}$, which represents the 3D coordinates of the $N$ vertices of the body surface, where $N=6890$. From this, we get a precise description of the human surface. Also, a global rotation $R\in \mathbb{R} ^{3\times 3}$, scale $s\in \mathbb{R} ^1$, and translation $t\in \mathbb{R} ^2$ can be obtained using camera parameters $\boldsymbol{\pi }$ based on weak perspective camera model. These three parameters are mainly used to project the 3D object onto the 2D image. The 3D object can be the human mesh vertices or 3D joints. Its specific usage will be described in detail in later sections.

\subsection{Feature Down-Sampling}\label{basemodel}
To facilitate understanding, the feature down-sampling of our work is first introduced with a single frame input as an example. 

As shown in Fig \ref{fig:mymodel}, we first input the image $I$ into the feature extractor without the last average pooling to get the feature $\boldsymbol{\boldsymbol{\phi} }_{0}^{s}\in \mathbb{R} ^{C_0\times W_0\times H_0}$. After that, the spatial features $\boldsymbol{\boldsymbol{\phi} }_{0}^{s}$ are fed into a set of deconvolutional networks $\left\{ DeConv_k \right\} _{k=1}^{3}$ to obtain $\left\{ \boldsymbol{\boldsymbol{\phi} }_{k}^{s}\in \mathbb{R} ^{C_k\times W_k\times H_k} \right\} _{k=1}^{3}$, i.e., 
\begin{equation}
    \boldsymbol{\boldsymbol{\phi} }_{k}^{s}=DeConv_k\left( \boldsymbol{\boldsymbol{\phi} }_{k-1}^{s} \right) ,\,\, \mathrm{for}\,\, k>0 . 
\end{equation}

Then we use the 2D projection $X_k$ of the obtained 3D human mesh vertices $M\left( \boldsymbol{\theta}_k ,\boldsymbol{\beta}_k \right)$ onto the feature map $\boldsymbol{\boldsymbol{\phi} }_{k}^{s}$ to obtain point-wise features $\boldsymbol{\boldsymbol{\phi} }_{k}^{p}\in \mathbb{R} ^{C_k}$, i.e.,
\begin{equation}
   \boldsymbol{\boldsymbol{\phi} }_{k}^{m}=\oplus \left\{ f\left( \boldsymbol{\boldsymbol{\phi} }_{k}^{p}\left( x_{k-1} \right) \right) ,\, \mathrm{for}\,\,x_{k-1}\,\,\mathrm{in} \, X_{k-1} \right\} ,\, k>1
\end{equation}
where $\oplus$ represents concatenation, $\boldsymbol{\boldsymbol{\phi} }_{k}^{p}\left( x_{k-1} \right)$ denotes acquiring $\boldsymbol{\boldsymbol{\phi} }_{k}^{p}$ according to $x_{k-1}$ using bilinear sampling, and $f\left( \cdot \right)$ is the MLP that reduces the channel dimension from $C_k$ to $C_m$. Then we get the feature $\boldsymbol{\boldsymbol{\phi} }_{k}^{m}\in \mathbb{R} ^{C_m*\tilde{N}}$, where $\tilde{N}$ is the number of mesh vertices. 

When $k=0$, it is worth noting that the information density of $\boldsymbol{\boldsymbol{\phi} }_{0}^{s}$ is very high. As illustrated in  Fig \ref{fig:process}, the 2D projection of the initial human mesh $\Theta_0$ obviously does not match the actual human body area. Performing projection down-sampling on $\boldsymbol{\boldsymbol{\phi} }_{1}^{s}$ thus cannot help the model to focus more on the human body area. In addition, the global information of the image is crucial to estimate the camera parameters. So we choose the grid sampling method to extract global features, when $k=1$. Grid sampling is that we define a $21\times 21$ grid to acquire point-wise features $\boldsymbol{\boldsymbol{\phi} }_{1}^{p}$. The other steps are the same as projection down-sampling. 

As for how $X_k$ is obtained, you can refer to this formula
\begin{equation}
    X_k=\Pi \left( \mathcal{D} \left( M\left( \boldsymbol{\theta } _k , \boldsymbol{\beta } _k \right) \right) \right) ,\,\, \mathrm{for}\,\, k>1 ,
\end{equation}
where $\Pi$ is an orthographic projection function based on camera parameters $\boldsymbol{\pi } _k$, and $\mathcal{D} \left( \cdot \right)$ represents down-sampling $\tilde{N}$ vertices from $N$ human mesh vertices. 

\subsection{Spatio-Temporal Alignment Fusion}
\subsubsection{Temporal Coherence Fusion Module}\label{moca}
\begin{figure}[t]
    \centering
    \includegraphics[width=0.99\linewidth]{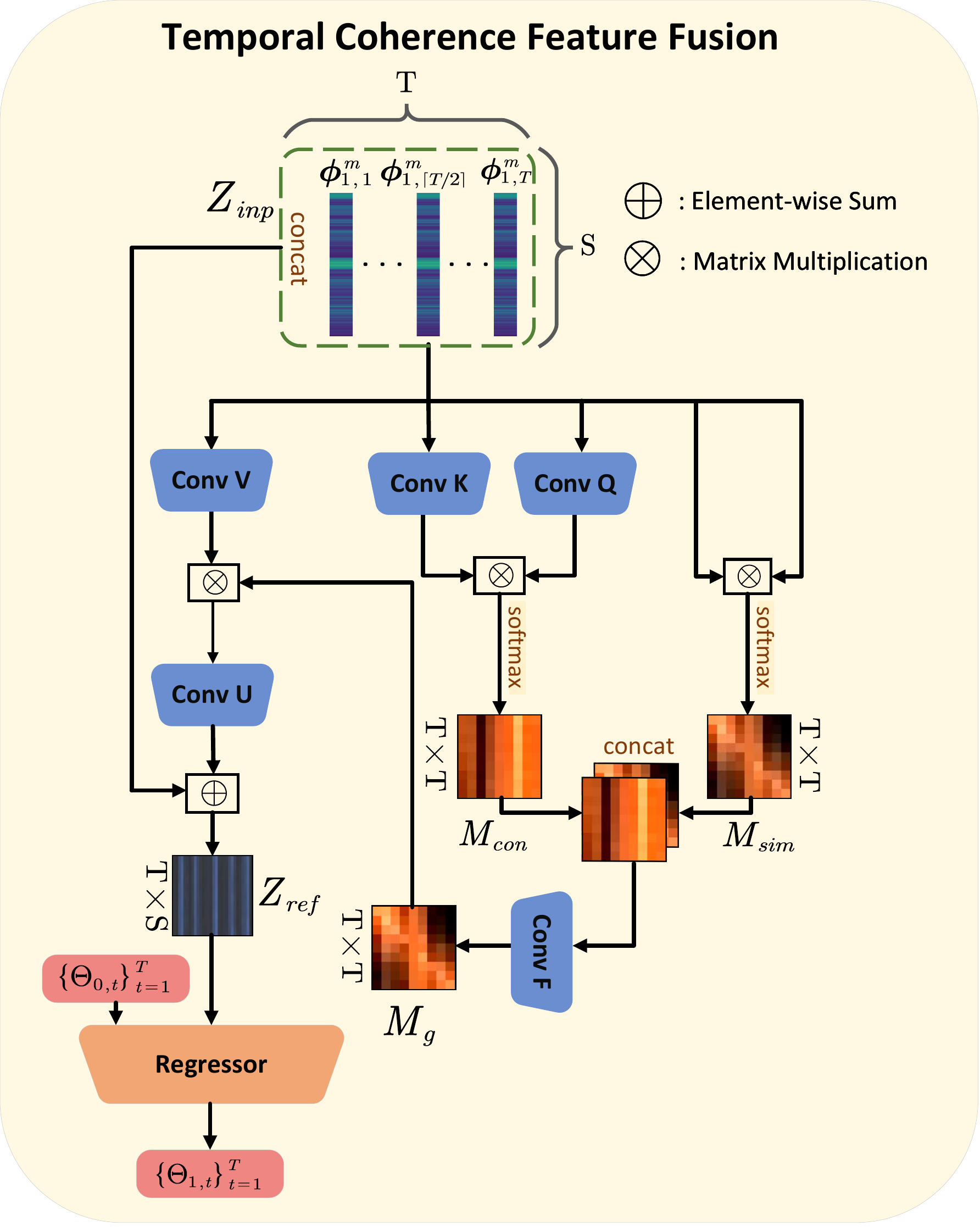}
    \caption{The structure of the temporal coherence fusion module. With $T$ features as input, the module outputs $T$ temporal refined features. We use TCFM to get initial human meshes. Note $\left\{ \Theta _{0,t} \right\} _{t=1}^{T}$ is set as the mean $\overline{\Theta }$ following \cite{hmr}. As for the correlation matrix, it calculates the coherence between the frames by multiplying two feature matrices. The correlation matrix is a $T \times T$ matrix. The element of the i-th row and j-th column represent the coherence between the i-th frame and the j-th frame. Larger values indicate stronger coherence. The brighter color indicates a larger value.}
    \label{fig:moca}
\end{figure}

For video-based models, an important design is how to implement feature interaction to capture temporal coherence effectively. Inspired by the non-local module in ~\cite{nonlocal} and \cite{wei2022capturing}, we introduce a lightweight temporal coherence fusion module, as illustrated in Fig \ref{fig:moca}. TCFM is a further improvement on the commonly used transformer structure. The main difference is that we add an extra correlation matrix $M_{sim}$. The traditional transformer usually encodes the features before computing the correlation matrix, as we get $M_{con}$ in Fig. \ref{fig:moca}. However, from the visualization of $M_{con}$, the network does not correctly establish the temporal coherence. The traditional transformer does not work as expected but only focuses on some frames with more information. Therefore, we additionally add $M_{sim}$ for steering the model so that each frame is more dependent on frames closer to itself and less on frames further away. As shown by $M_{sim}$ and $M_g$ in Fig. \ref{fig:moca}, the diagonal region is brighter, meaning TCFM learns the temporal coherence between frames more efficiently than traditional transformer.

The interaction objects of our temporal coherence fusion module are $\left\{ \boldsymbol{\phi }_{1, t}^m \right\} _{t=1}^T$, where $T$ is the number of input frames. As shown in the Fig \ref{fig:moca}, the feature matrix $Z_{inp}\in \mathbb{R} ^{T\times S}$ composed of $\left\{ \boldsymbol{\phi }_{1, t}^m \right\} _{t=1}^T$ is input to the module, where $S$ is the feature length of $\boldsymbol{\phi }_{1, t}^m$. We first input $Z_{inp}$ to three convolutional networks $\mathcal{Q}$, $\mathcal{K}$, $\mathcal{V}$ to obtain refined feature matrices $\left\{ Z_q,\,Z_k,\,Z_v \right\} \in \mathbb{R} ^{T\times \frac{S}{m}}$. After that, we obtain two correlation matrices
\begin{equation}
    \begin{cases}
    	M_{con}=softmax\left( Z_qZ_{k}^{T} \right)\\
    	M_{sim}=softmax\left( Z_{inp}Z_{inp}^{T} \right)\\
    \end{cases}
    \in \mathbb{R} ^{T\times T}.
\end{equation}
From the visualization of the correlation matrix $M_{con}$ in Fig \ref{fig:moca}, we can see that each frame's features are almost equally similar to the other frames', which is clearly not intuitive. Theoretically, every single frame should be more similar to the frames closer to itself. Therefore, for better feature refinement, higher weights should be given to more similar frames in the correlation matrix. In our work, in addition to $M_{con}$, we further use $M_{sim}$ to guide temporal coherence learning, i.e.,
\begin{equation}
    M_g=softmax\left( \mathcal{F} \left( concat\left( M_{con},\, M_{sim} \right) \right) \right),
\end{equation}
where $\mathcal{F}\left( \cdot \right)$ is a CNN that make $concat\left( M_{con}, M_{sim} \right) \in \mathbb{R} ^{2\times T\times T}$ downscale to $\mathbb{R} ^{T\times T}$. This module effectively enhances the ability of the model to learn long-range temporal features. Finally, we get the refined features $Z_{ref}$ by the following formula
\begin{equation}
    Z_{ref}=Z_{inp}+\mathcal{U} \left( M_g Z_{v} \right),
\end{equation}
where $\mathcal{U}\left( \cdot \right)$ is a convolutional layer, which let $M_g Z_{v} \in \mathbb{R} ^{T\times \frac{S}{m}}$ upscale to $\mathbb{R} ^{T\times S}$. With that, we can use the residual connection as in previous works. After that, we divide $Z_{ref}$ into $T$ features and feed them into the regressor separately to obtain a set of initial body meshes. These initial body meshes will be used to obtain spatial alignment clues and human spatial information for the next module SAFM.

\subsubsection{Spatial Alignment Fusion Module}\label{hafi}
\begin{figure}[ht]
    \centering
    \includegraphics[width=0.99\linewidth]{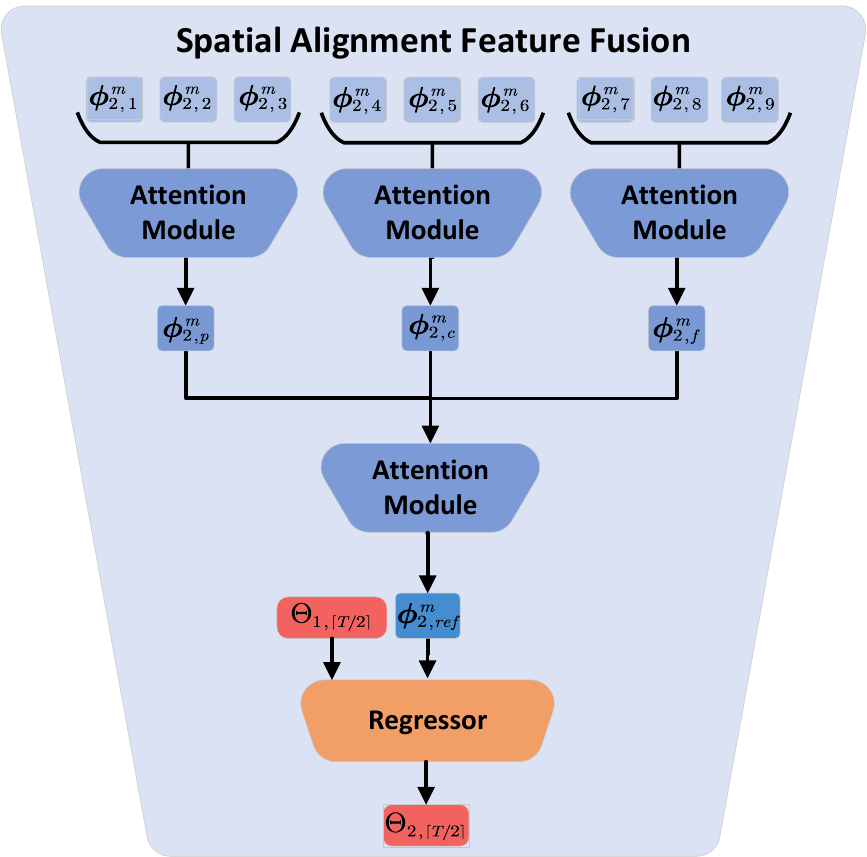}
    \caption{The structure of spatial alignment feature fusion module. Take the example of entering nine features $\left\{ \boldsymbol{\phi }_{2,1}^{m},\,\boldsymbol{\phi }_{2,2}^{m},\,\cdots ,\,\boldsymbol{\phi }_{2,9}^{m} \right\}$. Start with a group of three features and integrate them into one feature through the attention module. Then the three integrated features $\left\{ \boldsymbol{\phi }_{2,p}^{m},\,\boldsymbol{\phi }_{2,c}^{m},\,\boldsymbol{\phi }_{2,f}^{m} \right\}$ are integrated again into one feature $\boldsymbol{\phi }_{2,ref}^{m}$. We use $\boldsymbol{\phi }_{2,ref}^{m}$ to recover the 3D human mesh of the target frame.}
    \label{fig:hafi}
\end{figure}

Traditional video-based models often stop at exploiting temporal information. To overcome this issue, we propose SAFM to utilize the spatial information of each frame. "Spatial" is reflected in the fact that we do not directly use the full image information as input but further filter the spatial pixel alignment information for fusion. The most significant difference between spatial information and the previous temporal information is also reflected here. The focus we consider when fusing the temporal information is the temporal coherence of the input frames, so we take the full image information as input. However, in the second stage, we need more fine-grained information, i.e., spatial features of the human body. The pose of the human body tends to be different from frame to frame, but the shape is kept consistent. Meanwhile, the human body poses in neighboring frames tend to have some correlation. Based on the above discussion, we design a unique spatial feature fusion approach. SAFM can thus enhance the feature representation of the target frame and obtain more accurate recovery results. The structure of SAFM is shown in Fig \ref{fig:hafi}.

We illustrate how this module works with an input sequence of 9 features $\left\{ \boldsymbol{\phi }_{2,t}^{m} \right\} _{t=1}^{9}$, as we do in the final version of STAF. Following \cite{wei2022capturing}, we set each group to contain three frames, which has been shown to be the most efficient. As shown in Fig \ref{fig:hafi}, the feature sequences $\left\{ \boldsymbol{\phi }_{2,1}^{m},\,\boldsymbol{\phi }_{2,2}^{m},\,\cdots ,\,\boldsymbol{\phi }_{2,9}^{m} \right\}$ are first divided into three groups $\left\{ \boldsymbol{\phi }_{2,t-1}^{m},\,\boldsymbol{\phi }_{2,t}^{m},\,\boldsymbol{\phi }_{2,t+1}^{m} \right\}$, where $t=2,\,5,\,8$. We then input each group into an attention module to obtain the integrated features $\left\{ \boldsymbol{\phi }_{2,p}^{m},\,\boldsymbol{\phi }_{2,c}^{m},\,\boldsymbol{\phi }_{2,f}^{m} \right\}$. $\left\{ \boldsymbol{\phi }_{2,p}^{m},\,\boldsymbol{\phi }_{2,c}^{m},\,\boldsymbol{\phi }_{2,f}^{m} \right\}$ represent features of the \textbf{p}ast, \textbf{c}urrent and \textbf{f}uture frames, respectively. After that, we feed $\left\{ \boldsymbol{\phi }_{2,p}^{m},\,\boldsymbol{\phi }_{2,c}^{m},\,\boldsymbol{\phi }_{2,f}^{m} \right\}$ into the attention module again to get the final refined feature $\boldsymbol{\phi }_{2,ref}^{m}$ for the target frame. Note that all the attention modules mentioned above share the same network architecture and weights when deployed in practice.

Next, we describe how the attention module works. With $\left\{ \boldsymbol{\phi }_{2,t-1}^{m},\,\boldsymbol{\phi }_{2,t}^{m},\, \boldsymbol{\phi }_{2,t+1}^{m} \right\}$ as input, the attention module first reduces the dimension of each feature through a fully connected (FC) layer $fc\left(\cdot \right)$, i.e.,
\begin{equation}
    \boldsymbol{\phi }_{2,concat}^{m}=\oplus \left\{ fc\left( \boldsymbol{\phi }_{2,t-1}^{m} \right) ,\, fc\left( \boldsymbol{\phi }_{2,t}^{m} \right) ,\, fc\left( \,\boldsymbol{\phi }_{2,t+1}^{m} \right) \right\}, 
\end{equation}
where $\oplus$ represents concatenation. The resized feature $\boldsymbol{\phi }_{2,concat}^{m}$ is then passed through another three FC layers with tanh activation to reduce the channel size to $3$. We add a softmax activation in the end to calculate attention weights $\left\{ \alpha _1, \alpha _2, \alpha _3 \right\}$. Finally, we get the integrated feature
\begin{equation}
    \boldsymbol{\phi }_{2,integ}^{m}=\alpha_1\boldsymbol{\phi}_{2,t-1}^{m}+\alpha_2\boldsymbol{\phi}_{2,t}^{m}+\alpha_3\boldsymbol{\phi }_{2,t+1}^{m},\,t=2,\,5,\,8
\end{equation}
where $\boldsymbol{\phi }_{2,integ}^{m}$ are the features $\left\{ \boldsymbol{\phi }_{2,p}^{m},\,\boldsymbol{\phi }_{2,c}^{m},\,\boldsymbol{\phi }_{2,f}^{m} \right\}$ mentioned above. For example, $\boldsymbol{\phi }_{2,integ}^{m}$ is $\boldsymbol{\phi }_{2,p}^{m}$ when $t=2$.

 We use the features themselves to obtain attention weights, and then apply the attention weights to compute a weighted sum of the original features. This attention module fully preserves the spatial information of the original features, allowing this design to be embedded into other models without destroying the feature space. And this module can effectively tell the model which frame should be biased to integrate features better. It is worth mentioning that this multi-level integration approach considers only adjacent frames for each integration. Without such a multi-level design, it would get difficult to establish long-range spatial dependency. More importantly, model sizes would expand dramatically when the input sequence length gets too long. By adopting such a multi-level integration mechanism, SAFM can accommodate various input lengths.

\subsection{The Overall Model}\label{STAF}
At the end of this section, we present the overall structure of STAF, as shown in Fig \ref{fig:mymodel}. Given a sequence of images $\left\{ I_t \right\} _{t=1}^{T}$, a set of spatial features $\left\{ \boldsymbol{\phi }_{0,t}^{s}\in \mathbb{R} ^{C_0\times W_0\times H_0} \right\} _{t=1}^{T}$ is obtained after a CNN-based encoder. We mark the target frame with red borders in Fig \ref{fig:mymodel}. Then comes an essential operation, i.e.,
\begin{equation}
     \boldsymbol{\phi }_{0,\lfloor T/2 \rfloor}^{s\prime}=Avg\left( \left\{ \boldsymbol{\phi }_{0,t}^{s} \right\} _{t=1}^{T} \right), 
\end{equation}
where $Avg$ means average pooling. This module APM enables the model to rely less on the feature of the target frame $I_{\lceil T/2 \rceil}$ but take full advantage of the information of each frame. The feature obtained after average pooling is used to replace the original feature of the target frame. For convenience, $\boldsymbol{\phi }_{0,\lfloor T/2 \rfloor}^{s\prime}$ continues to be named $\boldsymbol{\phi }_{0,\lfloor T/2 \rfloor}^{s}$.

As described in Fig \ref{fig:mymodel}, the features $\left\{ \boldsymbol{\phi }_{0,t}^{s}\right\} _{t=1}^{T}$ are fed into the deconvolution network to get features $\left\{ \boldsymbol{\phi }_{1,t}^{s}\right\} _{t=1}^{T}$. And then $\left\{ \boldsymbol{\phi }_{1,t}^{s}\right\} _{t=1}^{T}$ are sampled by the grid to obtain the features $\left\{ \boldsymbol{\phi }_{1,t}^{m}\right\} _{t=1}^{T}$. Before sending the features into the regressor, we first feed them into the temporal coherence fusion module to fully learn the motion continuity dependencies. For more details, please refer to Sec \ref{moca}. This allows STAF to achieve not only better initial mesh recovery $\left\{ \Theta _{1,t} \right\} _{t=1}^{T}$, but also more accurate projection sampling used in the next step.

The features $\left\{ \boldsymbol{\phi }_{1,t}^{s}\right\} _{t=1}^{T}$ continue to be fed into the decoder consisting of deconvolution to obtain the feature sequence $\left\{ \boldsymbol{\phi }_{2,t}^{s}\right\} _{t=1}^{T}$. Unlike the former step, for the features $\left\{ \boldsymbol{\phi }_{2,t}^{m}\right\} _{t=1}^{T}$ obtained by projection sampling, we input them into the spatial alignment fusion module to obtain the feature $ \boldsymbol{\phi }_{2,ref}^{m}$ for the target frame. Owing to further deconvolution and projection sampling, the features $\left\{ \boldsymbol{\phi }_{2,t}^{m}\right\} _{t=1}^{T}$ contain rich fine-grained local information. The operation of multi-level adjacency integration can effectively enhance the mesh-alignment cues and enrich the human body information of $ \boldsymbol{\phi }_{2,\lceil T/2 \rceil}^{m}$. Finally, we feed $ \boldsymbol{\phi }_{2,ref}^{m}$ into the regressor together with the SMPL parameters $\Theta _{1,\lceil T/2 \rceil}$ obtained in the previous step to get the recovery result $\Theta _{2,\lfloor T/2 \rfloor}$ of the target frame.

For the last update of the SMPL parameters, we first send the features $ \boldsymbol{\phi }_{2,\lceil T/2 \rceil}^{s}$ into the decoder to get the features $ \boldsymbol{\phi }_{3,\lceil T/2 \rceil}^{s}$. Then we apply projection down-sampling to it to get the features $ \boldsymbol{\phi }_{3,\lceil T/2 \rceil}^{m}$. Finally, $ \boldsymbol{\phi }_{3,\lceil T/2 \rceil}^{m}$ concatenated with SMPL parameters $\Theta _{2,\lceil T/2 \rceil}$ is passed through the regressor to get the final result $\Theta _{3,\lceil T/2 \rceil}$.

\subsection{Loss Function}
For model training, we use three basic loss functions within the 3D human mesh recovery domain. Following TCMR~\cite{tcmr}, the first is the loss function $L_{smpl}$ of the SMPL parameters. It calculates the $L2$ loss between the predicted and ground-truth SMPL parameters. It should be noted that the datasets with the ground-truth SMPL parameters are very scarce. In order to take the vast datasets with ground-truth 2D and 3D joint coordinates into consideration, we introduce the other loss functions $L_{2D}$ and $L_{3D}$. The 3D joint coordinates can be obtained directly from the SMPL parameters, i.e., $X\left( \boldsymbol{\theta },\, \boldsymbol{\beta } \right) \in \mathbb{R} ^{3\times P}$, where $P$ is the number of joints. For the 2D joint coordinates $x$, we adopt the projection of 3D joints as follows:
\begin{equation}
    x=s\Pi \left( RX\left( \boldsymbol{\theta },\, \boldsymbol{\beta } \right) \right) +t
\end{equation}
where $\Pi$ is a projection function and $R$, $s$, $t$ are obtained from camera parameters. In conclusion, our loss function can be summarized as
\begin{equation}
    L=\lambda _{smpl}\left\| \Theta -\hat{\Theta} \right\| _2+\lambda _{3D}\left\| X-\hat{X} \right\| _2+\lambda _{2D}\left\| x-\hat{x} \right\| _2
\end{equation}
where $\lambda_{smpl}$, $\lambda_{3D}$ and $\lambda_{2D}$ are weights and would be 0 when relevant annotation is unavailable.

\begin{figure*}[h]
    \centering
    \includegraphics[width=0.99\textwidth]{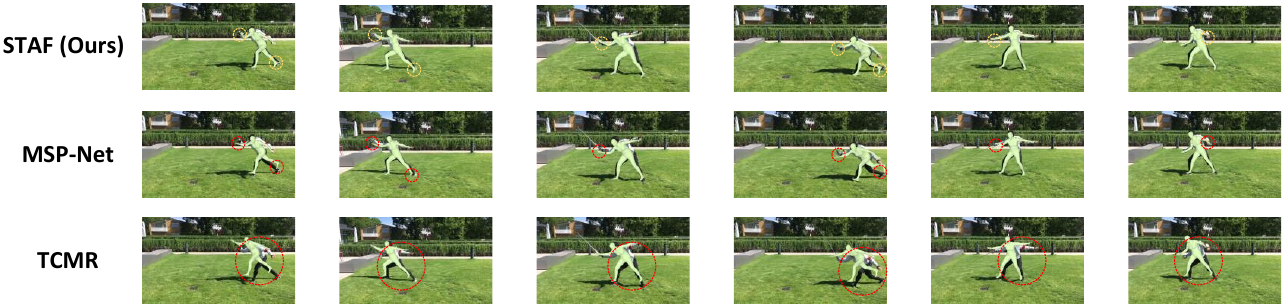}
    \caption{Qualitative comparison between STAF and two latest works (MPS-Net~\cite{wei2022capturing} and TCMR~\cite{tcmr}) . Traditional video-based models usually pursue only temporal coherence but miss spatial information, which might result in misalignment between the recovered mesh and image. Our STAF instead can effectively solve this problem.}
    \label{fig:process}
\end{figure*}

\begin{figure}[h]
    \centering
    \includegraphics[width=0.99\linewidth]{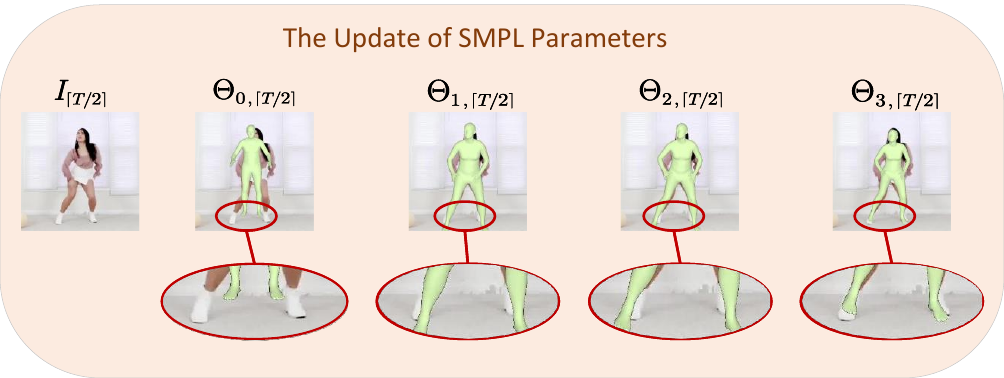}
    \caption{Visualization of output results of STAF regressors for each stage. It shows how the final results are obtained from SMPL mean parameters after adjustment by the three regressors.}
    \label{fig:process}
\end{figure}

\section{Experiments}
In this section, we describe the implementation details and experimental results. A series of experimental demonstrations and visualization results are also reported to prove the validity of the innovative points in our work.
\subsection{Datasets}
\textbf{COCO} Common Objects in Context \cite{coco} is a large-scale image dataset widely used for various computer vision tasks such as object detection, image segmentation, and image captioning. It is provided by Microsoft and consists of over 330,000 images, each with detailed annotations. For our task, we primarily utilize the part of the COCO that focuses on human subjects. COCO provides 2D joint location labels. And based on this, we use EFT~\cite{eft} to add pseudo labels, such as 3D joint positions and SMPL parameters, to the dataset. Since COCO is not a video dataset, we use it to train our base model only, allowing the base model to acquire the initial ability to extract features about the human body.

\textbf{LSP \& LSP-Extended} Leeds Sports Pose \cite{lsp} is a classic benchmark dataset used for human pose estimation. It consists of training and test sets, each containing 1,000 images with 2D joint labels. Later, LSP-Extended \cite{hahah} is introduced, which adds an additional 10,000 images for training. We only use the training set of LSP and LSP-Extended for training our base model. Additionally, we utilize pseudo-labels generated by EFT to enhance supervision.

\textbf{Human 3.6M}As a widely used 3D human body dataset, Human 3.6M~\cite{human36m} has been used as a benchmark dataset by many works for its large data volume and rich 3D labels. It should be noted that this dataset is collected indoors. It is thus often used together with in-the-wild datasets. However, its abundant 3D labels, stable objects and scenes are all useful to get human prior. Following previous work, we use subsets S1, S5, S6, S7, S8 for training and S9, S11 for testing. Since the videos in Human 3.6M are at 50 fps, which causes data redundancy, we extract frames at a frame rate of 25 fps. Note that the SMPL parameters obtained from Mosh are no longer publicly available due to legal reasons. Therefore, we use the pseudo SMPL labels provided by NeuralAnnot~\cite{moon2022neuralannot} to supervise the training following \cite{wei2022capturing} and \cite{tcmr}.

\textbf{MPII} MPII (Max Planck Institute for Informatics)~\cite{mpii} is a large-scale image dataset used for human pose estimation. This dataset is provided by the Max Planck Institute for Informatics in Germany and contains approximately 25,000 images along with corresponding pose annotations. We only use the images with complete 2D joint labels for training.

\textbf{3DPW} 3D Human Pose in the Wild \cite{3dpw} is a challenging dataset, since its data is collected from both indoors and outdoors. This dataset provides 3D joint coordinates, so we use it to enhance the model's adaptability to complex situations. Also, because it is very challenging, it is the main dataset for our experimental evaluation. We test both models trained with and without 3DPW to demonstrate the generalization ability of STAF.

\textbf{MPII3D} MPI-INF-3DHP~\cite{MPII3d} is also a dataset with 3D joints coordinates. It acquires ground-truth labels through a multi-camera marker-less motion capture system. It includes data obtained from indoors and outdoors, which is also a very tough dataset. And more and more works use it to perform experimental evaluation. In our experiment, we use MPII3D for both training and testing.

\textbf{Insta} InstaVariety~\cite{kanazawa2019learning} is a very large dataset with 2D labels, although its 2D joint coordinates are pseudo-labels generated by OpenPose. Its videos are collected from Instagram, so it is very content-rich and can complement the shortage of other datasets. We use it to perform weakly supervised training and enhance the generalization ability of the model.

\textbf{PoseTrack} PoseTrack~\cite{andriluka2018posetrack} is a multi-person video-based dataset with 2D labels. Although it is intended to provide a benchmark for pose estimation and multi-person tracking, we use it for training to increase the amount of training data.

Due to the two-stage training process of our model, the datasets used in each stage are not entirely the same. In the first stage, we train the base model with single-frame inputs, allowing us to utilize some nonvideo datasets. Following \cite{kocabas2021pare}, we use COCO~\cite{coco}, LSP~\cite{lsp}, LSP-Extended~\cite{hahah}, Human 3.6M~\cite{human36m}, MPII~\cite{mpii}, and MPII3D~\cite{MPII3d}. In the second training stage, we begin training the complete version of STAF, which requires video datasets. In addition to the previously mentioned Human 3.6M and MPII3D, we also incorporate 3DPW~\cite{3dpw}, Insta~\cite{kanazawa2019learning}, and PoseTrack~\cite{andriluka2018posetrack} for training, aiming to complement the limited training data. Overall, our training data volume remains consistent with previous works. Following previous works, we evaluate our approach in 3 classic benchmarks, i.e., 3DPW, MPII3D and Human 3.6M.

\subsection{Implementation Details}
We choose Resnet50~\cite{resnet} without the last average pooling as the encoder, which takes 9 images as input. It is worth mentioning that, in order to recover human meshes for all frames of a video, we choose to use a repeated set of 9 images as input for the first 4 frames and the final 4 frames. Since the image size is $224 \times 224$, the size of initial spatial features $\left\{\boldsymbol{\phi }_{0,t}^{s}  \right\} _{t=1}^{9}$  is $2048\times7\times7$. As for $\left\{ \boldsymbol{\phi }_{1,t}^{s},\, \boldsymbol{\phi }_{2,t}^{s},\, \boldsymbol{\phi }_{3,t}^{s} \right\} _{t=1}^{9}$, we keep their channel length constant, but their width and height are $\left\{ 14\times14,\, 28\times28,\, 56\times56 \right\}$. 

For the first regressor, since we use $21\times21$ grid sampling and further reduce the channel length from $C_k=2048$ to $C_m=5$, the size of the input features $\left\{ \boldsymbol{\phi }_{1,t}^{m}\right\} _{t=1}^{9}$ gets $21\times21\times5=2205$. For the other two regressors, we adopt projection down-sampling to calculate the features. Since the standard SMPL model generates too many vertices (6890), it is impracticable to use all of them to perform projection down-sampling. Following \cite{lin2021mesh}, we down-sample 6890 vertices to get a sparse human body mesh with only 431 vertices. The length of input features thus becomes $431\times5=2155$. To summarize, the features  $\left\{ \boldsymbol{\phi }_{1,t}^{m},\, \boldsymbol{\phi }_{2,t}^{m},\, \boldsymbol{\phi }_{3,t}^{m} \right\} _{t=1}^{9}$ have lengths of $\left\{ 2205,\, 2155,\, 2155 \right\}$, resp.

In the classical HMR~\cite{hmr} regressor, the input features are 2048 in length and go through three loops. Our regressors are consistent with the classical one but change the input scale. Considering that the HMR regressor takes three loops, we adopt a total of three regressors, too. Finally, in TCFM, the three convolutional networks $\mathcal{Q} ,\, \mathcal{K} ,\, \mathcal{V}$ reduce the input dimension 2205 by half to 1102.

\begin{table*}[h]
\setlength{\tabcolsep}{0.2mm}{
\begin{tabular}{clcccccccc}
\hline
                              & \multirow{2}{*}{Model} & \multirow{2}{*}{Backbone} & \multicolumn{4}{c}{3DPW}         & \multicolumn{3}{c}{MPI-INF-3D} \\\cline{4-10}
&&& PA-MPJPE $\downarrow$ & MPJPE $\downarrow$ & PVE $\downarrow$   & Accel $\downarrow$ & PA-MPJPE $\downarrow$   & MPJPE $\downarrow$   & Accel $\downarrow$   \\ \hline
\multirow{8}{*}{\rotatebox{90}{image-based}}  
& HMR~\cite{hmr} 2018          & ResNet-50     & 76.7     & 130.0 & -              & 37.4  & 89.8    & 124.2   & -       \\
& GraphCMR~\cite{kolotouros2019convolutional} 2019     & ResNet-50     & 70.2     & -     & -              & -     & -       & -       & -       \\
& SPIN~\cite{spin} 2019         & ResNet-50     & 59.2     & 96.9  & 116.4          & 29.8   & 67.5  & 105.2   &         \\
& I2L-MeshNet~\cite{moon2020i2l} 2020  & ResNet-50     & 57.7     & 93.2  & 110.1          & -      & -     & -       & -       \\
& PyMAF~\cite{pymaf} 2021        & ResNet-50     & 58.9     & 92.8  & 110.1          & -      & -     & -       & -       \\
& PARE \cite{kocabas2021pare} 2021         & HRNet-W32     & 50.9     & 82.0  & 97.9           & -      & -     & -       & -       \\ \hline
\multirow{11}{*}{\rotatebox{90}{video-based}} 
& HMMR \cite{kanazawa2019learning} 2019               & ResNet-50                 & 72.6     & 116.5 & 139.3 & 15.2  & -          & -       & -       \\
& Sim2Real \cite{doersch2019sim2real} 2019           & ResNet-50                 & 74.7     & -     & -     & -     & -          & -       & -       \\
& Temporal Context \cite{arnab2019exploiting} 2019   & ResNet-50 (from HMR)       & 72.2     & -     & -     & -     & -          & -       & -       \\
& DSD-SATN \cite{dsd} 2019           & ResNet-50                 & 69.5     & -     & -     & -     & -          & -       & -       \\
& MEVA* \cite{meva} 2020               & ResNet-50 (from SPIN)     & 54.7     & 86.9  & -     & 11.6  & 65.4       & 96.4    & 11.1    \\
& VIBE* \cite{vibe} 2020               & ResNet-50 (from SPIN)     & 51.9     & 82.9  & 99.1  & 23.4  & 64.6       & 96.6    & -       \\
& VIBE \cite{vibe} 2020               & ResNet-50 (from SPIN)     & 56.5     & 93.5  & 113.4 & 27.1  & 63.4       & 97.7    & -       \\
& TCMR* \cite{tcmr} 2021               & ResNet-50 (from SPIN)     & 52.7     & 86.5  & 102.9 & 7.1   & 63.5       & 97.3    & 8.5     \\
& TCMR \cite{tcmr} 2021               & ResNet-50 (from SPIN)     & 55.8     & 95.0  & 111.3 & \textbf{6.7}   & 62.8       & 97.4    & \textbf{8.0}     \\
& MPS-Net* \cite{wei2022capturing} 2022            & ResNet-50 (from SPIN)     & 52.1     & 84.3  & 99.7  & 7.4   & 62.8       & 96.7    & 9.6     \\
& MPS-Net \cite{wei2022capturing} 2022            & ResNet-50 (from SPIN)     & 54.0     & 91.6  & 109.6 & 7.5   & -          & -       & -       \\ \hline
& Ours*                   & ResNet-50 (pre-trained)    & \textbf{48.0}     & \textbf{80.6}  & \textbf{95.3}  & 8.2   & 59.6       & 93.7    & 10.0    \\
& Ours                   & ResNet-50 (pre-trained)    & 48.7     & 81.2  & 96.0  & 8.2   & \textbf{58.8}       & \textbf{92.4}    & 10.1    \\ \hline
\end{tabular}}
\caption{Comparison with SOTA Methods on 3DPW and MPII3D (* indicates training with 3DPW)}
\label{SOTA}
\end{table*}

Our base model consists of an encoder, three decoders, a down-sampling network and three regressors. It serves as our baseline for validating the effectiveness of the proposed modules. Additionally, a pre-trained base model is also utilized to provide a good initialization for STAF.

\subsection{Training Details}
\subsubsection{Stage 1} The base model is first trained on COCO~\cite{lin2014microsoft} for 175 epochs with a batchsize of 64. In the second stage, we train the base model on a mixed dataset for 60 epochs. Pseudo SMPL labels produced by EFT~\cite{eft} are used for supervising. The mixed dataset consists of Human 3.6M($50\%$), and MPII3D($20\%$). And the remaining $30\%$ of the mixed dataset is composed of COCO, LSP, LSP-Extended and MPII. The whole process takes about 4 days. 
 
 \subsubsection{Stage 2} When training STAF in our work, we use the pre-trained base model to initialize the parameters, except for the two modules TCFM and SAFM. Next, following \cite{meva,vibe,tcmr,wei2022capturing}, we train the network on a mixed dataset consisting of Insta, PoseTrack, Human 3.6M, 3DPW and MPII3D for 45 epochs with a mini-batchsize of 32. There are only $60\%$ of the training data with 2D labels. Note that Resnet50 is frozen during this stage of training. Image preprocessing including cropping method is referenced from VIBE~\cite{vibe} and MEVA~\cite{meva}. The training and testing video frame rate is 25 to 30 frames per second. Note that no data augmentation is applied in our work. The model weights are updated by the Adam optimizer with an initial learning rate of 0.00005. And the learning rate is reduced by a factor of 10 when the best performance is not updated for every 5 epochs. We train the model until it converges. In practical training, it typically takes around 18 hours.
 
 All training is performed on a single RTX 3090. The code implementation relies on Pytorch~\cite{paszke2017automatic}.

\subsection{Comparison with SOTA Methods}

To demonstrate the superiority of STAF, we first show its evaluation results on 3DPW, MPII3D, and Human 3.6M. We compare our model with other previous excellent models. The results are shown in Table \ref{SOTA} and Table \ref{h36m}. Following \cite{meva}, \cite{vibe}, \cite{tcmr}, \cite{wei2022capturing}, we use four standard evaluation metrics. The most comprehensive and representative metric is the mean per joint position error (MPJPE). Another important metric is PA-MPJPE, which expresses the Procrustes-aligned mean per joint position error. It removes the error introduced by the camera model by forcing the alignment. Note that PA-MPJPE evaluates only the accuracy of the recovered joints. The Per Vertex Position Error (PVE) calculates the error of the mesh vertices, but it is so redundant that it often does not match the actual qualitative result of the model. The units of the above metrics are all in $\mathrm{mm}$. Another key metric is the acceleration error (Accel), which is calculated as the acceleration error of the joint points in $\mathrm{mm}/\mathrm{s}^2$. It can be used to evaluate the smoothness of the reconstructed meshes. Note that the above joints and vertices are all in three dimensions.

We begin in Table \ref{SOTA} by summarizing the performance of some outstanding works over the past three years on 3DPW and MPII3D. These two datasets are chosen because they contain challenging in-the-wild data. The performance on these two challenging datasets can better demonstrate the model’s robustness. As seen from Table \ref{SOTA}, the all-around performance of STAF exceeds that of many previous SOTA models. STAF achieves optimal performance on three key metrics: PA-MPJPE, MPJPE, and PVE.  Compared with the latest work MPS-Net, STAF reduces the MPJPE by 3.7 $\mathrm{mm}$ and 3.0 $\mathrm{mm}$ on 3DPW and MPII3D, respectively. As mentioned earlier, in the past, it often has to sacrifice smoothness for precision, as in VIBE~\cite {vibe}, or conversely, sacrifice precision for smoothness, as in MEVA~\cite{meva}. STAF instead achieves a better trade-off between precision and smoothness. Our acceleration error remains very low while we achieve high reconstruction precision. In terms of smoothness, STAF far exceeds image-based models and is second only to TCMR and MPS-Net among video-based models. In Fig \ref{fig:accel}, we randomly select a video to test and plot the acceleration error. As shown, our model avoids severe jitter suffered by traditional video-based models and reaches a new level of overall smoothness. 

In addition to this, STAF shows surprising generalizability. The * in Table \ref{SOTA} indicates that the 3DPW training set is used for training, and the absence of * indicates that it is not used. From Table \ref{SOTA}, we can see that the PA-MPJPE and MPJPE of the previous models on 3DPW increase by 3.65-5.88\% and 8.7-12.8\%, respectively, when the models are not trained with the 3DPW training set. However, the PA-MPJPE and MPJPE of STAF increase only by 1.5\% and 0.7\%. On one hand, this can be explained by the small percentage of 3DPW in our training set, which accounts for only 0.5\%. On the other hand, it also demonstrates the stronger generalization ability of STAF, which can still achieve good evaluation results even without in-domain training data.

\begin{table}[h]
\setlength{\tabcolsep}{4.2mm}{
\begin{tabular}{ccc}
\hline
Model & \#Parameters (M) & Model Size (MB) \\ \hline
\rowcolor[HTML]{EFEFEF} 
VIBE  & 72.43            & 776             \\
MEVA  & 85.72            & 858.8           \\
\rowcolor[HTML]{EFEFEF} 
TCMR  & 108.89           & 1073            \\
Ours  & \textbf{51.12}            & \textbf{359.8}           \\ \hline
\end{tabular}}
\caption{Comparison of network parameters and model size.}
\label{parameter}
\end{table}

In order to further demonstrate the complexity and efficiency of STAF, we report the number of parameters and the model size of STAF compared to some other models in Table \ref{parameter}. However, due to our input consisting of only 9 frames, direct comparisons of FLOPs with models that utilize 16-frame inputs may not be entirely fair. Therefore, we disregard FLOPs in our comparison. From the perspective of parameters and model size, our model is significantly smaller than models that employ RNN or CNN to learn temporal information. Therefore, STAF exhibits higher model efficiency.

\begin{table}[h]
\setlength{\tabcolsep}{2.5mm}{
\begin{tabular}{cccc}
\hline
\multirow{2}{*}{Model} & \multicolumn{3}{c}{Human 3.6M} \\ \cline{2-4} 
                       & PA-MPJPE $\downarrow$   & MPJPE $\downarrow$   & Accel $\downarrow$   \\ \hline
HMMR \cite{kanazawa2019learning}                   & 56.9       & -       & -       \\
VIBE \cite{vibe}                  & 53.3       & 78.0    & 27.3    \\
MEVA \cite{meva}                   & 53.2       & 76.0    & 15.3    \\
TCMR \cite{tcmr}                   & 52.0       & 73.6    & 3.9     \\
MPS-Net \cite{wei2022capturing}               & 47.4       & \textbf{69.4}    & \textbf{3.6}     \\ \hline
Ours                   & \textbf{44.5}       & 70.4    & 4.8     \\ \hline
\end{tabular}}
\caption{Comparison with SOTA Methods on H36M}
\label{h36m}
\end{table}

Since H36M \cite{human36m} no longer publicly provides ground-truth SMPL parameters from Mosh, it is not fair to compare STAF directly with those models that use SMPL parameters from H36M for training. Contrary to the common perception, H36M is not an "easy" dataset, although its data are collected indoors. As mentioned in EFT \cite{eft}, many models that perform exceptionally well on H36M but poorly on 3DPW are often overfitting on the H36M training set. Therefore, we follow TCMR \cite{tcmr} and MPS-Net \cite{wei2022capturing} and reproduce some models without using the ground-truth SMPL parameters of H36M. The evaluation results are summarized in Table \ref{h36m}, where some of the results are from \cite{tcmr} and \cite{wei2022capturing}. It can be seen that STAF still achieves competitive results with equivalent training sets. STAF reduces PA-MPJPE by 2.9 $\mathrm{mm}$ compared to MPS-Net, which indicates STAF produces more precise human meshes.

\subsection{Ablation Study}
In this section, we demonstrate the contribution of our work. First, we validate the effectiveness of each module added to the base model. Then, we verify the applicability of APM to other models. Finally, we show how we determine the optimal way to combine TCFM and SAFM.
\begin{table}[t]
\setlength{\tabcolsep}{0.5mm}{
\begin{tabular}{ccccc}
\hline
                        & \multicolumn{4}{c}{3DPW}                                                            \\ \cline{2-5} 
\multirow{-2}{*}{model} & PA-MPJPE $\downarrow$         & MPJPE $\downarrow$    & PVE $\downarrow$  & Accel $\downarrow$        \\ \hline
base                   & 49.3                          & 83.5                  & 99.5              & 27.5         \\
base+APM               & 48.4                          & 81.8                  & 96.9              & \textbf{8.1}          \\
base+STAF              & 48.8                          & 82.3                  & 97.8              & 24.7         \\
base+APM+STAF   & \textbf{48.0}                        & \textbf{80.6} & \textbf{95.3} & 8.2 \\ \hline
\end{tabular}}
\caption{Ablation Results on 3DPW}
\label{ablation}
\end{table}

\subsubsection{Ablation Experiments}
We conduct a series of experiments on 3DPW to show the contribution of each module of STAF. The results are summarized in Tabel \ref{ablation}. In the table, APM denotes the average pooling module, STAF represents the combination of TCFM and SAFM, and base indicates the base model. For a fair comparison, the base model is also trained with the same second training stage as the subsequent ablation experiments. The evaluation results of the base model indicate that the acceleration error is still high, although the precision of human mesh recovery reaches a high level. This is a pain point that is difficult to be solved by many image-based models. Even many video-based models cannot improve the smoothness much. With the addition of the average pooling module, the acceleration error is easily reduced by 70.5\%, but the precision is not affected too much and even increased. STAF also brings an all-round improvement, with PA-MPJPE, MPJPE, PVE, and Accel reduced by 1.3 $\mathrm{mm}$, 1.2 $\mathrm{mm}$, 4.2 $\mathrm{mm}$ and 2.8 $\mathrm{mm/s^2}$, respectively. From the last row, we can see that the combination of APM and STAF achieves the best results. Although the acceleration error increases by 0.1 $\mathrm{mm/s^2}$ compared to base+APM, the precision is improved. A good balance between precision and smoothness is achieved.

\begin{table}[h]
\setlength{\tabcolsep}{1.3mm}{
\begin{tabular}{ccccc}
\hline
                                                            & \multicolumn{4}{c}{3DPW}                                      \\ \cline{2-5} 
\multirow{-2}{*}{Model}                                     & PA-MPJPE $\downarrow$      & MPJPE $\downarrow$         & PVE $\downarrow$            & Accel $\downarrow$        \\ \hline
HMR                                                         & 54.3          & 91.4          & 107.2          & 29.1         \\
\rowcolor[HTML]{EFEFEF}
HMR+HAFI                                                     & 54.2 & 90.9          & 107.3          & 29.1          \\
HMR+APM                                                     & 53.1 & 87.9          & 103.3          & \textbf{7.7}          \\
\rowcolor[HTML]{EFEFEF}
\begin{tabular}[c]{@{}c@{}}MPS-Net\\ w/o HAFI\end{tabular}      & 53.0          & \textbf{86.7}          & \textbf{102.2}          & 23.5         \\
\begin{tabular}[c]{@{}c@{}}MPS-Net\\ w/ HAFI\end{tabular} & 53.2          & 87.7          & 102.9          & 8.0          \\
\rowcolor[HTML]{EFEFEF}
\begin{tabular}[c]{@{}c@{}}MPS-Net\\ w/ APM\end{tabular}  & \textbf{52.8} & 87.5 & 102.7 & 7.8 \\ \hline
\end{tabular}}
\caption{Ablation Results of Average Pooling Module on 3DPW}
\label{avg}
\end{table}
\subsubsection{Effect of Average Pooling Module}
Next, we demonstrate the effect of our average pooling module, and the related results are in Table \ref{avg}. Our inspiration is drawn from the HAFI module of MPS-Net~\cite{wei2022capturing}. The evaluation results of MPS-Net w/o HAFI are from \cite{wei2022capturing}. The evaluation results of MPS-Net w/ HAFI are reproduced by ourselves and are similar to the results of \cite{wei2022capturing}. It can be found that MPS-Net achieves such a low acceleration error relying mainly on the HAFI module.

However, we do not achieve the same effect when adding HAFI to the classic model HMR~\cite{hmr}. Since the output of HAFI is a weighted sum of the features, we output the weights obtained from both the pre-trained MPS-Net and HMR+HAFI. Note that HMR+HAFI represents taking a sequence as input to HAFI and then sending the integrated features to HMR. As shown in Fig \ref{fig:attention}, compared to HMR+HAFI, MPS-Net does not focus on the target frame effectively but on the whole input sequence.

Obviously, HMR+HAFI is more reasonable since our goal is to get the result for the target frame, which typically is the middle frame of the input sequence. So, the middle weight deserves to be the largest. However, our experiment results demonstrate that it is the over-reliance on the feature of the target frame that leads to high acceleration error. A similar point has been mentioned in TCMR~\cite{tcmr}. The evaluation results of MPS-Net w/ APM and MPS-Net w/ HAFI in Table \ref{avg} also prove our point. Next, we replace the HAFI module of MPS-Net with our APM and find the acceleration error is still reduced. Therefore, we can conclude that HAFI's ability to minimize the acceleration error sharply is not attributed to its attention module design but benefits from the equal treatment of each frame, i.e., attaching similar weights to features of each frame. Another weak point of HAFI is its poor generalization ability. In most cases, its attention module is still automatically biased toward the target frame during training. Our APM, instead, is easier to generalize because it forces the model to handle each frame equally. We also test it on the classical model HMR~\cite{hmr} to verify its generalizability. The effect is noticeable, with a 73.5\% drop in acceleration error. Hence, we believe APM is a simple and effective way to improve smoothness and can be easily embedded in both image-based and seq2frame video-based models.

\begin{table}[t]
\setlength{\tabcolsep}{0.8mm}{
\begin{tabular}{lcccc}
\hline
\multirow{2}{*}{Model} & \multicolumn{4}{c}{3DPW}                                                                                       \\ \cline{2-5} 
                       & \multicolumn{1}{l}{PA-MPJPE $\downarrow$} & \multicolumn{1}{l}{MPJPE $\downarrow$} & \multicolumn{1}{l}{PVE $\downarrow$} & \multicolumn{1}{l}{Accel $\downarrow$} \\ \hline
TCFM1                  & 48.5                         & 81.8                      & 97.1                    & 8.1                       \\
TCFM1+SAFM1            & 49.4                         & 83.2                      & 98.4                    & 8.5                       \\
TCFM1+SAFM2            & \textbf{48.0}                & \textbf{80.6}             & \textbf{95.3}           & 8.2                       \\
TCFM1+SAFM3            & 48.8                         & 81.6                      & 96.6                    & 8.3                       \\
TCFM2                  & 48.9                         & 82.7                      & 98.5                    & 8.2                       \\
TCFM2+SAFM2            & 48.8                         & 81.8                      & 97.4                    & 8.3                       \\
TCFM2+SAFM3            & 48.9                         & 82.2                      & 97.6                    & 8.4                       \\
TCFM3                  & 49.2                         & 83.2                      & 99.1                    & \textbf{8.0}              \\
TCFM3+SAFM3            & 49.0                         & 82.0                      & 97.3                    & 8.1                       \\
SAFM1                  & 48.4                         & 81.3                      & 96.2                    & 8.2                       \\
SAFM2                  & 48.5                         & 80.7                      & 95.7                    & 8.2                       \\
SAFM3                  & 48.8                         & 82.1                      & 97.0                    & \textbf{8.0}              \\ \hline
\end{tabular}}
\caption{Ablation Results of Spatio-Temporal Fusion Module on 3DPW}
\label{tf}
\end{table}

\subsubsection{Ablation Study of TCFM and SAFM}
\label{TFII}
The final ablation experiment is to find the best combination of TCFM and SAFM. First, we introduce the meaning of the first column in Table \ref{tf}. TCFM refers to \textbf{T}emporal \textbf{C}oherence \textbf{F}usion \textbf{M}odule and SAFM refers to \textbf{S}patial \textbf{A}lignment \textbf{F}usion \textbf{M}odule. And the $n$ in TCFM$n$ and SAFM$n$ indicates that the input features of this module are $\left\{ \boldsymbol{\phi }_{n,t}^{m} \right\} _{t=1}^{9}$. Note that the output feature sequence of TCFM$n$ is the input of SAFM$m$, when $n=m$. Because SAFM must come after TCFM, and to avoid bloat, we do not consider module reuse. So there are 4+3+2+3=12 combinations in total. The average pooling module is also applied during this experiment.

\begin{figure}[h]
    \centering
    \includegraphics[width=0.99\linewidth]{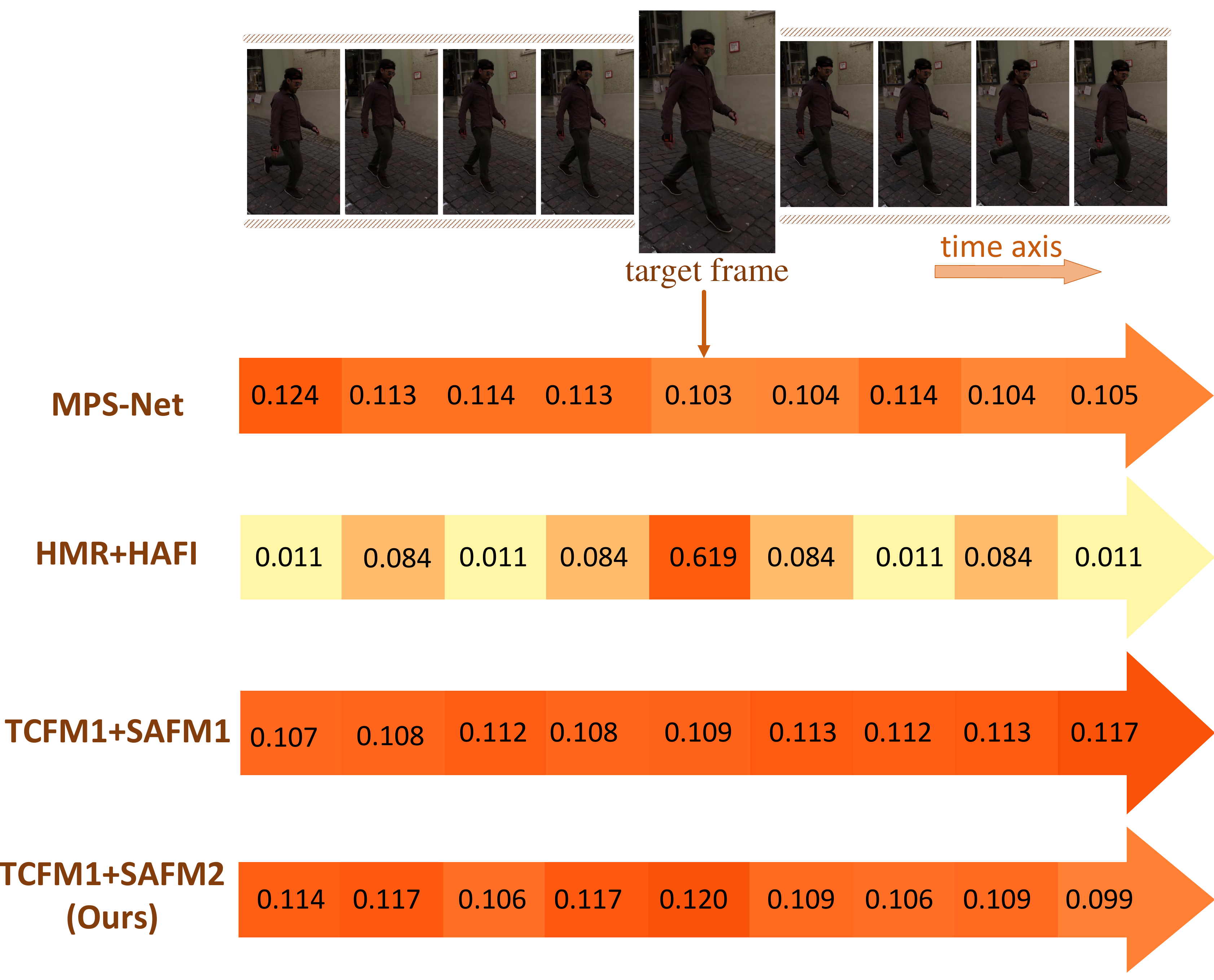}
    \caption{The attention weights generated by the attention module of each model. MPS-Net is \cite{wei2022capturing}. HMR+HAFI refers to a seq2frame video-based model composed of a classical single-frame model \cite{hmr} plus HAFI. As for TCFM1+SAFM1 and TCFM1+SAFM2, please refer to Sec \ref{TFII}. As we can see, neither MPS-Net nor TCFM1+SAFM1 can focus on the target frame correctly. HMR+HAFI instead focuses too much on the target frame and cannot take into account the temporal coherence. Our STAF, however, can focus on the whole input sequence with a slight bias towards the target frame so as to obtain a better balance between precision and smoothness.}
    \label{fig:attention}
\end{figure}

The evaluation results of all combinations are presented in Table \ref{tf}. The best combination is finally found, i.e., TCFM1+SAFM2. As for why this is the case, it is explainable. First, in an iterative error feedback loop, the latter regressor outputs a smaller $\bigtriangleup \Theta $, i.e., the lower-level features have less impact on the final result. As shown in Fig \ref{fig:process}, the output of the first regressor is very close to the final result, and the last two regressors just need to do a little fine-tuning on the details. Therefore, the benefit of refining higher-level features is supposed to be greater. The evaluation results also prove this point. The evaluation metrics of TCFM1 to TCFM3 and SAFM1 to SAFM3 in Table \ref{tf} both show an increasing trend. To answer why TCFM1+SAFM2 is better than TCFM1+SAFM1, we output the weights generated by SAFM in TCFM1+SAFM2 and TCFM1+SAFM1. As shown in Fig \ref{fig:attention}, the attention weights generated by TCFM1+SAFM2 are in line with our expectation that the attention model is only slightly biased towards the middle frame. However, the attention weights generated by TCFM1+SAFM1 are obviously unreasonable.

We believe that this is because SAFM1 adopts the refined feature of TCFM1 as input. But TCFM1 destroys the spatial structure of the original features, making SAFM1 difficult to learn them correctly. More importantly, if TCFM1+SAFM1 is used, SAFM cannot well use the human spatial information as well as mesh-alignment cues of each frame to enhance the feature representation of the target frame. TCFM1+SAFM1 is similar to traditional video-based models because they all just apply a temporal encoding of the features. Although TCFM1+SAFM2 is not the lowest in acceleration error, it is optimal in all other evaluation metrics. So, it is finally chosen under comprehensive consideration.

\begin{figure}[th]
    \centering
    \includegraphics[width=0.99\linewidth]{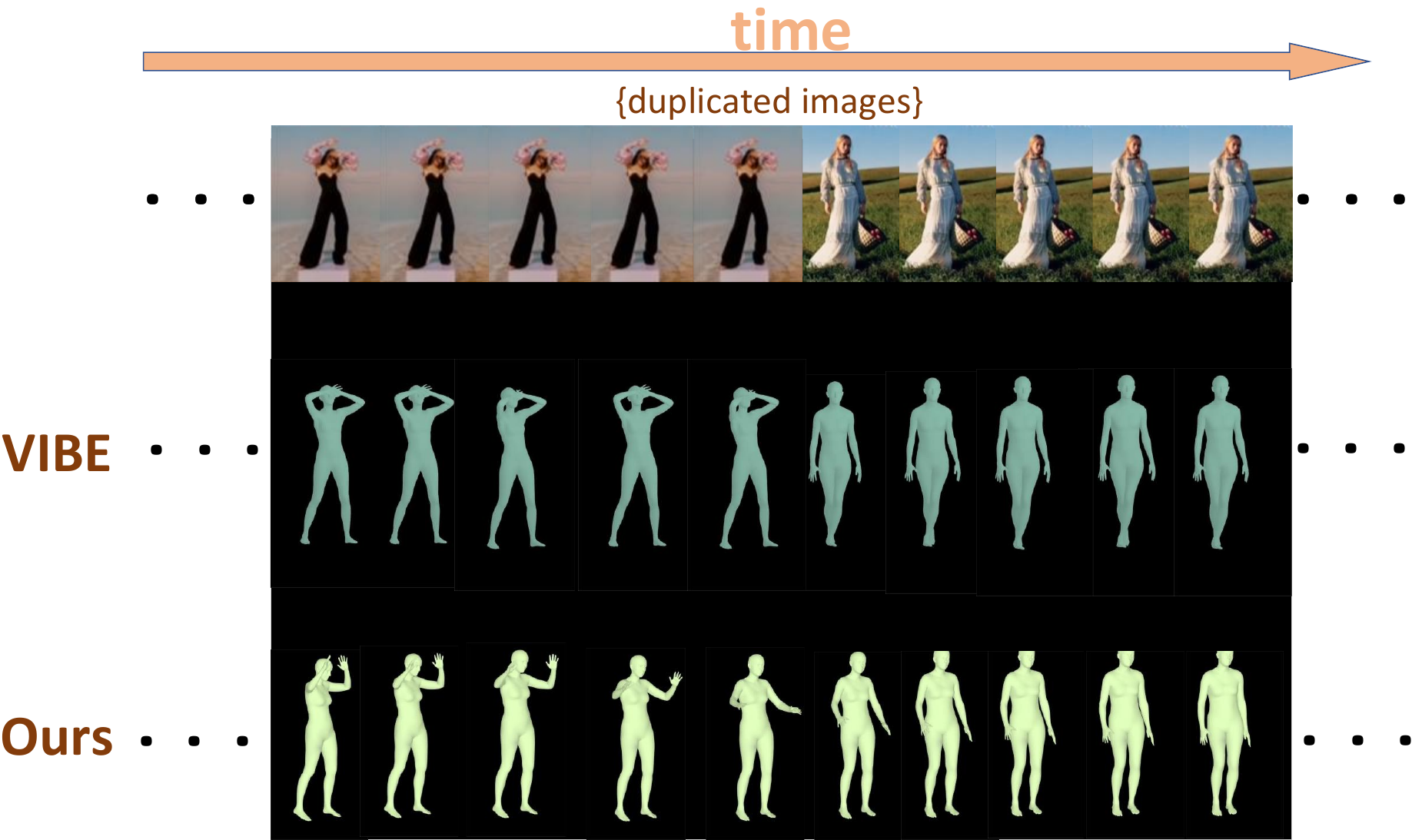}
    \caption{Visualization of an extreme example, where the human pose in the video suddenly changes dramatically. Compared to VIBE~\cite{vibe}, STAF can estimate a smoother human motion process.}
    \label{fig:smooth}
\end{figure}

\section{Discussion}
In this section, we would like to discuss the issue of over-smooth in STAF. As can be seen in Fig \ref{fig:smooth}, we design an extreme example in which we forcefully merge two individuals with different poses into a single video as input. While models like VIBE~\cite{vibe}, which prioritize accuracy, generate poses without any transition, STAF generates smooth transitions from one pose to another. On one hand, this demonstrates that our model does indeed exhibit over-smooth in such extreme cases. On the other hand, this example also showcases the capability of STAF to estimate smooth results.

To address this issue, we adopt a shorter input sequence in our model. This is done to prevent an excessive sequence length, which could impact the precision of recovery from the target frame. It is also essential to be aware that if sequences are too brief, it may be challenging to acquire enough temporal information.

Taking these factors holistically into account, we choose to use a 9-frame input sequence, striking a balance between smoothness and precision. For more qualitative results, please refer to our project page. We also encourage you to run our program to generate video demos.
\section{Conclusion}

In this paper, we presented a novel seq2frame video-based model for 3D human mesh recovery. We proposed spatio-temporal alignment fusion to preserve spatial information and further exploit both temporal and spatial information. We introduced the temporal coherence fusion module that takes full advantage of the motion coherence without destroying the original feature space. In addition to the temporal encoder, we proposed the spatial alignment fusion module. We cleverly used spatial information and alignment cues to further correct the recovery result of the target frame. Except for the above, we revealed the cause of the temporal discontinuity that previous works suffer from, i.e., over-reliance on the target frame. We thus proposed the averaging pooling module, which reduces the model's reliance on the target frame and enhances the overall attention of the input sequence. It improved the smoothness substantially without affecting the recovery precision and can be easily embedded in other image-based and seq2frame video-based models. Compared with the previous 3D human mesh recovery models, STAF achieved a better trade-off between precision and smoothness.
{
    \small
    \bibliographystyle{ieeenat_fullname}
    
}


\end{document}